\documentclass[conference]{IEEEtran}
\IEEEoverridecommandlockouts

\usepackage{pbalance}

\IEEEoverridecommandlockouts 
\usepackage[dvips]{graphicx}
\usepackage{algorithmic}
\usepackage{algorithm}

\usepackage{listings}
\usepackage{xcolor}
\usepackage[OT4,T1]{fontenc}

\usepackage[cmex10]{amsmath}
\usepackage{url}
\usepackage{multirow}

\usepackage{cite}
\usepackage{makecell}
\usepackage{stfloats}

\title{Gradient Boosting Trees and Large Language Models for Tabular Data Few-Shot Learning}
\author{
\IEEEauthorblockN{Carlos Huertas}
\IEEEauthorblockA{Amazon Research\\
Email: carlohue@amazon.com}

}

\begin{document}
\maketitle              

\begin{abstract}
Large Language Models (LLM) have brought numerous of new applications to Machine Learning (ML). In the context of tabular data (TD), recent studies show that TabLLM is a very powerful mechanism for few-shot-learning (FSL) applications, even if gradient boosting decisions trees (GBDT) have historically dominated the TD field. In this work we demonstrate that although LLMs are a viable alternative, the evidence suggests that baselines used to gauge performance can be improved. We replicated public benchmarks and our methodology improves LightGBM by 290\%, this is mainly driven by forcing node splitting with few samples, a critical step in FSL with GBDT. Our results show an advantage to TabLLM for 8 or fewer shots, but as the number of samples increases GBDT provides competitive performance at a fraction of runtime. For other real-life applications with vast number of samples, we found FSL still useful to improve model diversity, and when combined with ExtraTrees it provides strong resilience to overfitting, our proposal was validated in a ML competition setting ranking first place.
\end{abstract}

\section{Introduction}
%
%

Tabular data in real-world applications is the most common type of data\cite{ravid2022}, this continues to be true since relational databases are still pretty common in all sort of domains from social to natural sciences \cite{wang2017,yoon2020,zhang2020,wang2020,fang2024}. Deep Learning (DL), or in general, Neural Network based architectures have shown tremendous potential in tasks like Natural Language Processing (NLP) with developments like transformers\cite{vaswani2017} and large-scale pre-trained models like DeBERTa\cite{he2020} have pushed the state-of-the-art (SOTA) and gave DL a top spot in performance. The same can be observed for Computer Vision (CV) with developments like convolutional neural networks (CNN) opening the door for more advanced designs like EfficientNets\cite{tan2019} and more recently Vision Transformers (ViT) have found their way into CV as well\cite{alexey2021} with Next-ViT\cite{li2022} aiming to bridge the gap that still separates ViT from CNN in terms of efficiency in the latency/accuracy trade-off. 

Despite all the success from DL, tabular data continues to be omnipresent\cite{vadim2021,liu2022}, and to the best of our knowledge, we have not found a consistent DL-based approach that can outperform Gradient Boosted Decision Trees (GBDT) \cite{xgboostPaper, lgbmPaper,cbPaper} over a \textit{wide variety} of tasks and conditions, even though it is possible to find specific niche setups where this happens\cite{ziv2021,poggio2020,grin2022}.

Recently, the introduction of Large Language Models (LLM)\cite{naved2024} demonstrated a whole new level of performance for several tasks\cite{wei2023,buceck2023}, from traditional NLP to even code generation\cite{jiang2023}. The concept of revisiting the qualities of DL-based techniques, in particular LLM for tabular data surged again\cite{fang2024}, due to some of the key properties over GBDT\cite{good2016}, such as: representation learning, sequential processing and generalization. Even though DL provides some advantages, if maximum performance is desired, GBDT continues to be the SOTA\cite{borisov2024} even with amazing advances in DL, some of the most notable attempts to outperform GBDT with DL methods include: Wide\&Deep\cite{wideDeep}, DeepFM\cite{deepFM}, SDTR\cite{sdtr}, DeepGBM\cite{deepgbm}, TabNN\cite{tabnn}, BGNN\cite{ivanov2021}, TabNet\cite{tabnet2019}, TransTab\cite{transtab}, TabTransformer\cite{tabtrans}, SAINT\cite{saint} and NPT\cite{npt}, none of them providing enough evidence to actually be able to beat GBDT over a wide variety of tasks, most of the time, it has been demonstrated the claimed improvements are only present in very specific cases or datasets\cite{ziv2021}.

There are however, some situations where LLM based solutions seem to have an edge\cite{fang2024}, this is when data is limited, and LLM have the capacity to perform both zero-shot (ZSL) and few-shot learning (FSL)\cite{soysal2020}. While there is no doubt current SOTA in GBDT will show random-performance for zero-shot learning, recent studies\cite{unipredict} show that even under a few-shot schema, LLM can outperform Xgboost\cite{xgboostPaper}, one of the most popular GBDT algorithms.

In this work, we will further explore the performance of GBDT under a FSL schema in order to provide strong baselines. Since previous studies\cite{ziv2021} have demonstrated bias in claims of DL outperforming GBDT in other tasks, we look to enhance experiments to confirm SOTA results in the new trend of results regarding FSL and the superiority of LLM over GBDT.

\section{Related Work}

The main concept behind ZSL or FSL by definition implies the evaluated classifier has either (a) never seen the data samples before (ZSL), or only a few samples (FSL), however, this can only be proven true if we were to train a model (LLM for the purpose of this research) from scratch. Any sort of pre-trained architecture could, in theory, already seen the dataset, hence showing incredible performance. This particular problem has been studied before\cite{elephants2024}, where both GPT-3.5 and GPT-4 are proven to have seen common datasets in the past, like \textit{Adult Income} and \textit{FICO}\cite{ficodataset}, in some cases, even proven LLM have literally memorized the datasets verbatim\cite{carlini2021} as samples can be extracted out. With this in mind, the fair evaluation of LLM vs GBDT under a truly FSL schema is very challenging, while we can guarantee GBDT has never seen the data, the same cannot be said for many LLM applications.

The results from Bordt et al.\cite{elephants2024}, using a 20-shot-learning are shown in Table \ref{tableElephant}, in this work authors study LLM memorization. 

\begin{table*}[tbp]
\caption{GPT-3/4 vs Traditional Algorithms for Few-Shot-Learning Performance (AUC) }\label{tableElephant}
\centering
\begin{tabular}{|@{\vrule width0ptheight9pt\enspace}c|c|c|c|c|c|c|c|}\hline
\hfil
\bf Algorithm&\bf \makecell{Kaggle\\Titanic} &\bf \makecell{OpenML\\Diabetes} & \bf \makecell{Adult\\Income}  &\bf FICO&\bf \makecell{Spaceship\\Titanic} &\bf Pneumonia  \\\hline
GPT-4&\textbf{0.98}&0.75&0.82&0.68&0.69&0.81  \\\hline
GPT-3.5&0.82&0.74&0.79&0.65&0.63&0.54  \\\hline
GBDT (Xgboost)&0.84&0.75&\textbf{0.87}&\textbf{0.72}&\textbf{0.80}&\textbf{0.90} \\\hline
Logistic Regression &0.79&\textbf{0.78}&0.85&\textbf{0.72}&0.77&\textbf{0.90}\\\hline
\end{tabular}
\end{table*}

Although LLM results are far from bad, the performance still shows gaps to match GBDT. On top of this, GBDT is a much simpler and faster model, essentially being a more efficient and more powerful option. For the Kaggle Titanic dataset, the power of GPT-4 might look impressive, until authors have proven this is due to memorization and not any particular useful learning. This problem is not particular to tabular data, as LLM have been proved to do so as well for other domains\cite{carlini2023}. Nonetheless, authors have found that there is some learning happening, for datasets with no memorization LLM can still provide some performance, especially in very few shot-learning, which leads to the work of  Hegselmann et al.\cite{tabllm}, where LLM are shown to actually outperform GBDT.

In such work, authors present TabLLM, a very innovative solution to use LLMs for few-shot classification on tabular data, in principle, first running a serialization-step, to turn tabular into a natural language representation. An extensive analysis is done to benchmark multiple serialization techniques. Surprisingly, one of the simplest approaches resulted to be very effective, \textit{``Text Template''} is a compact representation of the form: \textit{``The <column name> is <value>''}. This followed by a task-specific prompt, that can later be fined-tuned for FSL.

TabLLM has been benchmarked for both binary and multi-class problems, from datasets identified in key literature for this task\cite{kadra2021,grin2022,borisov2024}. For simplicity, we will focus on the binary tasks, as to ensure all tasks are of the same objective, and metrics are comparable, e.g. AUC. A summary of their benchmarking results is presented in Table \ref{tabletabllm}. For full details refer to Table 12, 13 and 14 in \cite{tabllm}.

\begin{table*}[h]
	\caption{TabLLM Experiments Results: LightGBM (GBDT) vs NN-Based (AUC) }\label{tabletabllm}
	\centering
	\begin{tabular}{|@{\vrule width0ptheight9pt\enspace}c|c|c|c|c|c|c|c|c|}\hline
		\hfil
		\bf Dataset&\bf Method  & \bf 4-shot  &\bf 8-shot &\bf 16-shot  &\bf 32-shot  &\bf 64-shot &\bf Average  \\\hline
		
		Bank \cite{kadra2021} &\makecell{LightGBM\\ \bf TabPFN\\TabLLM}
		&\makecell{ 0.50 \\ 0.59 \\ 0.59 }
		&\makecell{ 0.50 \\ 0.66 \\ 0.64 }
		&\makecell{ 0.50 \\ 0.69 \\ 0.65 }
		&\makecell{ 0.50 \\ 0.76 \\ 0.64 } 
		& \makecell{ 0.77 \\ 0.82 \\ 0.69 } 
		& \makecell{ 0.554 \\ \bf 0.704 \\ 0.642 } \\\hline
		
		Blood \cite{kadra2021} &\makecell{LightGBM\\ \bf TabPFN\\TabLLM}
		&\makecell{ 0.50 \\ 0.52 \\ 0.58 }
		&\makecell{ 0.50 \\ 0.64 \\ 0.66 }
		&\makecell{ 0.50 \\ 0.67 \\ 0.66 }
		&\makecell{ 0.50 \\ 0.70 \\ 0.68 }
		& \makecell{ 0.69 \\ 0.73 \\ 0.68 }
		& \makecell{ 0.538 \\ \bf 0.652 \\ 0.652 } \\\hline
		
		Credit-g \cite{kadra2021} &\makecell{LightGBM\\TabPFN\\ \bf TabLLM}
		&\makecell{ 0.50 \\ 0.58 \\ 0.69 }
		&\makecell{ 0.50 \\ 0.59 \\ 0.66 }
		&\makecell{ 0.50 \\ 0.64 \\ 0.66 }
		&\makecell{ 0.50 \\ 0.69 \\ 0.72 }
		& \makecell{ 0.61 \\ 0.70 \\ 0.70 }
		& \makecell{ 0.522 \\ 0.640 \\  \bf 0.686 }  \\\hline
		
		Diabetes \cite{diabetesDS} &\makecell{LightGBM\\ \bf TabPFN\\TabLLM}
		&\makecell{ 0.50 \\ 0.61 \\ 0.61 }
		&\makecell{ 0.50 \\ 0.67 \\ 0.63 }
		&\makecell{ 0.50 \\ 0.71 \\ 0.69 }
		&\makecell{ 0.50 \\ 0.77 \\ 0.68 }
		& \makecell{ 0.79 \\ 0.82 \\ 0.73 }
		& \makecell{ 0.558 \\ \bf 0.716 \\ 0.668 }  \\\hline
		
		Heart \cite{HeartDS}&\makecell{LightGBM\\ \bf TabPFN\\TabLLM}
		&\makecell{ 0.50 \\ 0.84 \\ 0.76 }
		&\makecell{ 0.50 \\ 0.88 \\ 0.83 }
		&\makecell{ 0.50 \\ 0.87 \\ 0.87 }
		&\makecell{ 0.50 \\ 0.91 \\ 0.87 }
		& \makecell{ 0.91 \\ 0.92 \\ 0.91 } 
		& \makecell{ 0.582 \\ \bf 0.884 \\ 0.848 } \\\hline
		
		Income \cite{vadim2021} &\makecell{LightGBM\\TabPFN\\ \bf TabLLM}
		&\makecell{ 0.50 \\ 0.73 \\ 0.84 }
		&\makecell{ 0.50 \\ 0.71 \\ 0.84 }
		&\makecell{ 0.50 \\ 0.76 \\ 0.84 }
		&\makecell{ 0.50 \\ 0.80 \\ 0.84 }
		& \makecell{ 0.78 \\ 0.82 \\ 0.84 } 
		& \makecell{ 0.556 \\ 0.764 \\ \bf 0.840 } \\\hline
		
	\end{tabular}
\end{table*}

The results show NN-based solutions, both TabPFN\cite{TabPFN} and TabLLM\cite{tabllm}, substantially outperform LightGBM for FSL, the improved performance by these techniques is such that the minimal delta observed comes in the \textit{Bank} dataset where TabLLM shows an average (over 4 to 64 FSL) advantage of 163\% relative improvement  $[(0.642-0.5)/(0.554-0.5)]$ vs the GBDT solution. On the other extreme, the superiority of TabLLM goes further to outperform LightGBM for as much as 745\% $[(0.686-0.5)/(0.522-0.5)]$ for the \textit{Credit-g} experiment.

In the next section, we present our analysis regarding the extreme underperformance from LightGBM, and our recommendations to establish a fair baseline for a FSL application. Increasing its performance to a more competitive level, and hoping this serves as reference for future benchmarks in the field.

\section{Proposed Solution}

The process of FSL might have slightly different interpretations depending on the field, but the core concept remains, the usage of only a few samples to train a model. This concept holds for the tabular data use-case. Knowing this, is imperative to understand how algorithms like LightGBM work in order to build an effective FSL solution. The LightGBM algorithm is a boosting approach using decisions trees (DT) to learn a function from the input space $X^{s}$ to the gradient space $G$ \cite{lgbmPaper}, the splitting criteria is reviewed below.

Given a training set with $n$ i.i.d. instances $\{x_{1},...,x_{n}\}$ , where each $x_{i}$ is a vector with dimension $s$ in space $X^{s}$. For each boosting iteration, the negative gradients of the loss function with respect to the output of the model are denoted as $\{g_{1},..., g_{n}\}$ . The DT model splits each node to maximize information gain, which is measured by the variance \textit{after} splitting. For a training set $O$ on a fixed node, the gain of splitting ($V$) feature $j$ at point $d$ is defined as:

\begin{eqnarray}
	V_{j|O}(d)= \frac{1}{n_{O}} 
	\bigg( \frac{        \left(   \Sigma_{  \{     x_{i} \in:x_{ij}\le d \} }   g_{i}    \right)^2           }  {n_{l|O}^j (d)}        +  \frac{ \left(   \Sigma_{  \{     x_{i} \in:x_{ij}> d \} }   g_{i}    \right)^2  } {n_{r|O}^j (d)}
	\bigg)
\end{eqnarray}

The problem however arises in practice since the optimization is constrained so that the left $n_{l|O}^j (d)$ and right  $n_{r|O}^j (d)$ nodes have a minimum sample size. A segment of LightGBM implementation is shown in Algorithm \ref{algo:splitting}.

\begin{algorithm}[h]
\caption{LightGBM: feature\_histogram Implementation}

\begin{lstlisting}[language={[11]C++}]
is_splittable_ = false;

//...
const auto grad = GET_GRAD(data_, t);
const auto hess = GET_HESS(data_, t);

sum_left_gradient += grad;
sum_left_hessian += hess;

left_count += cnt;

if (left_count < min_data_in_leaf) {
	continue;
}

right_count = num_data - left_count;
if (right_count < min_data_in_leaf) {
	break;
}
//...

is_splittable_ = true;
\end{lstlisting}

\label{algo:splitting}
\end{algorithm}

The \textit{minimum samples per leaf} then becomes a blocker for FSL, causing the algorithm to stall. Unable to perform any split until training samples exceeds the \textit{min\_samples\_leaf} parameter. Although previous works\cite{tabllm} have explored parameter tuning based on literature recommendations\cite{vadim2021,grin2022}, this is not being addressed, and as a result LightGBM shows \textit{random-guess} performance (e.g. 0.5 AUC) in most experiments, since the default value for \textit{min\_samples\_leaf} is set to $20$. 

In this work we propose a LightGBM configuration specifically for FSL applications. We identified key parameters needed as shown in Table \ref{params}.

\begin{table*}[h] 
	\caption{Proposed Parameters for FSL Applications in LightGBM}\label{params}
	\centering
	\begin{tabular}{|@{\vrule width0ptheight9pt\enspace}c|c|c|c|}\hline
		\hfil
		\bf Parameter&\bf Description  & \bf Default & \bf Recommended   \\\hline
		
		extra\_trees & use extremely randomized trees
		& false & True \\\hline
		
		num\_leaves & max number of leaves in one tree
			& 31 & 4 \\\hline
		
		eta & shrinkage rate
			& 0.1 & 0.05 \\\hline
		
		\bf min\_data\_in\_leaf & \bf minimal number of data in one leaf
			& \bf 20 & \bf 1 \\\hline
		
		feature\_fraction & subset of features on each tree
			& 1.0 & 0.5 \\\hline
			
		bagging\_fraction & select part of data without resampling
		& 1.0 & 0.5 \\\hline
		
		bagging\_freq & frequency for bagging
		& 0 & 1 \\\hline
		
		min\_data\_per\_group & number of data per categorical group
		& 100 & 1 \\\hline
		
		cat\_l2 & L2 regularization in categorical split
		& 10 & 0 \\\hline
		
		cat\_smooth & reduce noise-effect in categoricals
		& 10 & 0 \\\hline
		
		max\_cat\_to\_onehot & one-vs-other algorithm control
		& 4 & 100 \\\hline
		
		min\_data\_in\_bin & minimal number of data inside one bin
		& 3 & 3 \\\hline

	\end{tabular}
\end{table*}

The most important parameter for FSL is, without a doubt, \textit{min\_data\_in\_leaf}, as otherwise optimization cannot happen.

The same concept applies to any other parameter that relies on counting of samples, such as \textit{min\_data\_per\_group}. In general, it is required to minimize the restrictions here, this is however a very bad practice for Non-FSL applications, leading to overfitting, and should be used with care in any other types of problems.

Due to the partition mechanism of DT, small sample-size will generate a very constrained histogram, and a greedy partition threshold is not desirable, to enhance this, the usage of extremely randomized trees is required to ensure partition splits are over represented in the tree structure.

In the next section we provide experimental results to demonstrate the ability of LightGBM to do few-shot learning.

\section{Experiments}

Our experiment design covers two folds. First, we replicate previous work \cite{tabllm}, but apply our recommended methodology to enable efficient FSL for LightGBM. Second, we bring a practical application to incorporate FSL into larger-scale data, this serves as reference that even if samples are vast, FSL can provide benefits.

\subsection{TabLLM Experiment Replication}

Both TabPFN and TabLLM show similar performance in average. Only a marginal improvement of 1\% in favor of TabPFN, however, both of those solutions outperform LightGBM over 343\% in average, with extreme cases such as Credit-g where the relative performance of TabLLM is 745\% better. While we were able to validate these numbers are correct, our results show this extreme underperformance is driven due to incorrect parameters.

We have replicated the binary problems. For the sake of simplicity, our LightGBM does not include hyperparameter tuning and instead executed with our fixed recommended parameters as shown in Table \ref{params}. This leads to intentional underoptimization to disregard the effect of better tuning in the results. We found LightGBM much more competitive as seen in Table  \ref{ourresults}.

Our methodology improved the performance of LightGBM by 290\%, essentially reducing both TabLLM and TabPFN claimed advantage by 84.5\%. 

LightGBM can outperform or meet TabLLM for 64-shot performance in 5 out of 6 datasets, only missing for Income dataset, where TabLLM performance is constant regardless the number of shots. This is an interesting problem to review for memorization. 

For extreme low FSL, like 4 and 8 shot, we found LightGBM to be competitive, yet falling generally behind, this can further be improved with parameter tuning, but gaps are large to close still. Over 16-shots there is considerable performance parity and as the shots increase LightGBM consistently starts to take over. When enough samples are available, no performance advantages were found from TabLLM or TabPFN, yet both solutions are considerably slower to LightGBM.

\begin{table*}[h]
	\caption{Untuned LightGBM Improved Baseline Performance (AUC) }\label{ourresults}
	\centering
	\begin{tabular}{|@{\vrule width0ptheight9pt\enspace}c|c|c|c|c|c|c|c|c|}\hline
		\hfil
		\bf Dataset&\bf Method  & \bf 4-shot  &\bf 8-shot &\bf 16-shot  &\bf 32-shot  &\bf 64-shot &\bf Average  \\\hline
		
		Bank &\makecell{  Our LightGBM \\ LightGBM \cite{tabllm} }
		&\makecell{ 0.54 \\ 0.50 }
		&\makecell{ 0.62 \\ 0.50  }
		&\makecell{ 0.65 \\ 0.50 }
		&\makecell{ 0.70 \\ 0.50  } 
		& \makecell{ 0.77 \\ 0.77 } 
		& \makecell{ \bf 0.656 \\ 0.554 } \\\hline
		
		Blood&\makecell{  Our LightGBM \\ LightGBM \cite{tabllm} }
		&\makecell{ 0.50 \\ 0.50 }
		&\makecell{ 0.63\\ 0.50 }
		&\makecell{ 0.67 \\ 0.50 }
		&\makecell{ 0.70 \\ 0.50 }
		& \makecell{ 0.71 \\ 0.69 }
		& \makecell{ \bf 0.642 \\ 0.538 } \\\hline
		
		Credit-g&\makecell{  Our LightGBM \\ LightGBM \cite{tabllm} }
		&\makecell{ 0.60 \\ 0.50 }
		&\makecell{ 0.64 \\ 0.50 }
		&\makecell{ 0.62\\ 0.50 }
		&\makecell{ 0.65 \\ 0.50 }
		& \makecell{ 0.70 \\ 0.61 }
		& \makecell{ \bf 0.642 \\ 0.522 } \\\hline
		
		Diabetes&\makecell{  Our LightGBM \\ LightGBM \cite{tabllm} }
		&\makecell{ 0.50 \\ 0.50 }
		&\makecell{ 0.62 \\ 0.50 }
		&\makecell{ 0.65 \\ 0.50 }
		&\makecell{ 0.71 \\ 0.50 }
		& \makecell{ 0.78 \\ 0.79 }
		& \makecell{ \bf 0.652 \\ 0.558 } \\\hline
		
		Heart&\makecell{  Our LightGBM \\ LightGBM \cite{tabllm} }
		&\makecell{ 0.78 \\ 0.50 }
		&\makecell{ 0.85\\ 0.50 }
		&\makecell{ 0.88 \\ 0.50 }
		&\makecell{ 0.90 \\ 0.50 }
		& \makecell{ 0.91 \\ 0.91 }
		& \makecell{ \bf 0.864 \\ 0.582 } \\\hline
		
		Income&\makecell{  Our LightGBM \\ LightGBM \cite{tabllm} }
		&\makecell{ 0.60 \\ 0.50 }
		&\makecell{ 0.68 \\ 0.50 }
		&\makecell{ 0.77 \\ 0.50 }
		&\makecell{ 0.81 \\ 0.50 }
		& \makecell{ 0.83 \\ 0.78 }
		& \makecell{ \bf 0.738 \\ 0.556 } \\\hline
		
	\end{tabular}
\end{table*}

\subsection{FedCSIS 2024 Data Science Challenge}

To further review performance and applications of FSL, we applied our findings to the FedCSIS 2024 Data Science Challenge hosted in the KnowledgePit platform, a web system designed for ML competitions helping to bring collaboration between industry and academia \cite{knowledgePit}.

The challenge: \textit{Predicting Stock Trends}, provides stock-tickers and their performance as measured by 116 financial-markers, such as: \textit{Dividend Payout Ratio}, \textit{Gross Profit Margin}, and \textit{Price to Total Revenue per Share}. The information is provided for current Trailing Twelve Months (TTM), these are static features, named \textit{I1} to \textit{I58}. Another set, known as relative-features, named \textit{dI1} to \textit{dI58} provide the relative 1-yr change for such indicators.

This is a competition event that promotes an objective evaluation of performance. Participants were asked to predict the optimal investment strategy of securities among 3 actions: \textit{buy}, \textit{hold} or \textit{sell}.  An in-depth review of the competition is detailed in \cite{fedcsis2024comp}.

\textbf{Initial Model:} In order to establish a baseline we started our simplest possible solution directly with DT, this due to its usual superiority over other algorithms for tabular data that has not been deeply feature engineered \cite{huertas2023}. A LightGBM regression model using all features as-is and the original discrete target \textit{``Class''} achieves 0.8439 mean absolute error (MAE). The first insight came from feature importance, which suggests the relative (\textit{dI*}) variables far dominate the static set (\textit{I*}) as seen in Table \ref{compFeatImp}, taking 4 out of the top 5 spots. This inspired further review to enhance generalization given the limited data size.

\begin{table*}[h] 
	\caption{Competition: Top Financial Indicators as determined by LGBM Baseline Model}\label{compFeatImp}
	\centering
	\begin{tabular}{|@{\vrule width0ptheight9pt\enspace}c|c|c|c|}\hline
		\hfil
		\bf Feature  & \bf Description & \bf Importance   \\\hline
		
	 	dI58 & 1-year Absolute Change of Price to Cash Flow from Operations per Share & 1.000 \\\hline
	 	I57 & Cash Flow from Operations Pct of Capital Expenditures & 0.725 \\\hline
	 	dI52 & 1-year Absolute Change of Cash Ratio & 0.675 \\\hline
	 	dI43 & 1-year Absolute Change of Dividend Yield - Common - Net - Issue - \% & 0.613 \\\hline
	 	dI56 & 1-year Absolute Change of Book Value Percentage of Market Capitalization & 0.537 \\\hline
	 	I5 & Excess Cash Margin - \% & 0.536 \\\hline
	 	dI57 & 1-year Absolute Change of Cash Flow from Operations Pct of Capital Expenditures & 0.521 \\\hline
	 	Group & Industry sector & 0.520 \\\hline
	 	I24 & Accounts Receivable Turnover & 0.471 \\\hline
	 	dI17 & 1-year Absolute Change of Debt - Total to EBITDA & 0.404 \\\hline
	 	dI44 & 1-year Absolute Change of PE Growth Ratio & 0.377 \\\hline
		
	\end{tabular}
\end{table*}

\textbf{Sample and Feature Selection:} Following Occam's Razor principle, we challenged the value of the static features (\textit{I*}). When using all variables it’s possible to get 0.6018 AUC, an alternate variant for diversification would be to use relative-features (\textit{dI*}) only, this proves to be quite competitive, retaining 95\% of predictive power (0.5963 AUC), with a 50\% reduction of features. This is important since the large feature mismatch promotes orthogonal decisions boundary for subsequent ensembling techniques.

Another diversification technique comes from instance sampling. We studied the sample-size vs performance in the same binary case to determine the right number of shots to use, ideally the smaller the better for diversification in further stages. Results are provided in Table \ref{compSampleSelect} where we can observe even after a 40\% sample size reduction (6864 to 4118) there is zero impact in performance, and reducing further brings minimal degradation, this provides an ideal framework for FSL, as the ability to use few samples allows for stacking level-0 models with non-overlapping samples.

\begin{table*}[h] 
	\caption{Competition: Sample Size Effect in Performance}\label{compSampleSelect}
	\centering
	\begin{tabular}{|@{\vrule width0ptheight9pt\enspace}c|c|c|c|}\hline
		\hfil
		\bf Sample Size  & \bf AUC  \\\hline
		
		6,864 & 0.6018 \\\hline
		
		6,178 & 0.6098 \\\hline
		
		5,491 & 0.6027 \\\hline
		
		4,118 & 0.6055 \\\hline
		
		1,373 & 0.5835 \\\hline
		
		686 & 0.5887 \\\hline

	\end{tabular}
\end{table*}

\textbf{Stacking Level-0 Models:} Based on previous insights, we determined that FSL is a viable strategy to enable multiple orthogonal models. Although previous analysis was done in a binary setting, these new models are built with the \textit{Perform} target in the dataset. Unlike the discrete buy/hold/sell, this continuous representation allows the model to understand the impact of each action, e.g. not all \textit{“buys”} are equal, since they provide different levels of financial gain/loss. Using a 3k shot-approach per model we forced diversification in the sample space. In order to improve generalization, we used the learnings that ExtraTrees outperforms GBDT in most FSL settings. We did not create any feature engineering, but our \textit{Base Feature Set} is a concatenation of existing features over multiple years for stock-tickers that are present more than once in the dataset, only relative features (\textit{dI*}) are used. The details of each model and their respective performance is shown in Table \ref{compeModels}. Note that because we switched to \textit{Perform} as target, MAE is no longer optimal, so we optimized for the mean squared error (MSE) instead.

\begin{table*}[h] 
	\caption{Level-0 Models for FedCSIS24: Stock Prediction Competition}\label{compeModels}
	\centering
	\begin{tabular}{|@{\vrule width0ptheight9pt\enspace}c|c|c|c|}\hline
		\hfil
		\bf Model &\bf Features  & \bf Target & \bf MSE   \\\hline
		
		ExtraTrees & Base Feature Set
		& Original & 0.020039 \\\hline
		
		GBDT & Base Feature Set
		& Original & 0.020051 \\\hline
		
		ExtraTrees & Base Feature Set
		& Quantile(0.5\%,99.5\%) & 0.019609 \\\hline
		
		ExtraTrees & Base with Categorical Removed
		& Original & 0.020088 \\\hline
		
		ExtraTrees & Base with static features added back
		& Original & 0.020055 \\\hline
		
	\end{tabular}
\end{table*}

\textbf{Final Blend:} Our Level-1 Meta model is fed with the five different L0 configurations. \textit{MLPRegressor} from \textit{sklearn} was selected for simplicity, architecture is 2 hidden-layers of 10 and 5 neurons with ReLU activation\cite{relu}. Optimization is still using \textit{Perform} target, with a 10\% validation sample size and adam optimizer\cite{adamOptimizer}. Early stopping is based on R2 score with 64 max epochs.

Since the competition requires discrete actions (buy/hold/sell) instead of expected performance, we optimize the performance-to-action thresholds by ensuring the same action-distribution between train and test. This solution has ranked $1^{st}$ place in the event, with a MAE score of 0.772, which represents a 3.66\% and 7.12\% relative improvement against $2^{nd}$  and $10^{th}$  place respectively.

\section{Conclusions}

When the merit of a proposal is measured by its relative performance to a baseline, the baseline itself is equally, or even more important than the proposal. It is trivial to show a solution is good by simply selecting a weak reference point to compare with. Efforts invested in a new proposal can also be applied to improve a baseline. In this work we have improved LightGBM FSL performance found in literature by 290\%. Improvements of this magnitude are unusual with just parameter optimization.

Our results show GBDT can perform few-shot-learning (FSL) with surprising performance with as little as 8-shots. And when data is available, FSL can be used to force diversification between individual models in ensemble or stacking architectures.

While global optimum is too expensive to reach, its imperative to learn the inner caveats of algorithms to exploit their strengths to reasonable levels. Our solution in FedCSIS competition shows the importance of understanding your algorithms to maximize performance, both the FSL approach for diversity and ExtraTrees to fight overfitting proved to be very successful in our experiments to achieve $1^{st}$ place.

\bibliographystyle{IEEEbib}
\bibliography{arxivGbdtLLM}

\end{document}